%% file: main.tex
\documentclass[11pt]{article}
\usepackage[preprint]{acl}
\usepackage{times}
\usepackage{latexsym}
\usepackage[T1]{fontenc}
\usepackage[utf8]{inputenc}
\usepackage{microtype}
\usepackage{inconsolata}
\usepackage{graphicx}
\usepackage{booktabs}
\usepackage{amsmath}
\usepackage{xcolor}
\usepackage{xurl}
\newcommand{\bestResolveOriginal}{29.75}
\newcommand{\chainFeedbackReported}{55.9}
\newcommand{\chainFeedbackWritten}{80.9}
\newcommand{\chainLift}{1.12}
\newcommand{\chainLiftHigh}{1.28}
\newcommand{\chainLiftLow}{1.04}
\newcommand{\deltaResolveMaxFullTwentyEight}{+2.78}
\newcommand{\deltaResolveMeanFullTwentyEight}{+0.47}
\newcommand{\deltaScoreMaxFullTwentyEight}{+2.89}
\newcommand{\deltaScoreMeanFullTwentyEight}{+0.84}
\newcommand{\deltaScoreMinFullTwentyEight}{-0.26}
\newcommand{\goZeroResolveMeanFull}{5.71}
\newcommand{\goZeroResolveMeanOfficial}{4.35}
\newcommand{\goZeroScoreMeanFull}{42.08}
\newcommand{\goZeroScoreMeanOfficial}{38.33}
\newcommand{\kappaCrossModelHigh}{0.75}
\newcommand{\kappaCrossModelLow}{0.50}
\newcommand{\kappaCrossModelMedian}{0.63}
\newcommand{\kappaDiffHigh}{0.40}
\newcommand{\kappaDiffLow}{-0.17}
\newcommand{\kappaHoldoutAnnotators}{0.92}
\newcommand{\kappaSameModelHigh}{0.94}
\newcommand{\kappaSameModelLow}{0.66}
\newcommand{\kappaSameModelMedian}{0.81}
\newcommand{\liftGroupI}{1.67}
\newcommand{\liftGroupIHigh}{2.04}
\newcommand{\liftGroupILow}{1.32}
\newcommand{\liftGroupIi}{1.08}
\newcommand{\liftGroupIiHigh}{1.41}
\newcommand{\liftGroupIiLow}{0.83}
\newcommand{\liftSameNoLeak}{1.66}
\newcommand{\matchedRatio}{1.55}
\newcommand{\matchedRatioHigh}{1.72}
\newcommand{\matchedRatioLow}{1.40}
\newcommand{\memorizationPartialP}{0.98}
\newcommand{\memorizationPartialR}{-0.01}
\newcommand{\memorizationWithinDiff}{-11.9}
\newcommand{\memorizationWithinP}{0.24}
\newcommand{\minutesColdAudit}{28.1}
\newcommand{\nAddendumConventions}{153}
\newcommand{\nAdjacentPairs}{34}
\newcommand{\nAllFailAOne}{50}
\newcommand{\nAllFailATwo}{14}
\newcommand{\nAllFailBehaviour}{17}
\newcommand{\nAllFailEnvironment}{646}
\newcommand{\nAllFailLiteral}{5}
\newcommand{\nAllFailNoBuild}{5{,}338}
\newcommand{\nAllFailTotal}{6{,}167}
\newcommand{\nAllFailUnexplained}{97}
\newcommand{\nBlockedTestsPTwoP}{2{,}333}
\newcommand{\nBlockedUnitTests}{3{,}191}
\newcommand{\nChainFlagged}{64}

\newcommand{\nChainSteps}{155}
\newcommand{\nConfirmedItemsChain}{28}
\newcommand{\nConfirmedItemsDeepSwe}{1}
\newcommand{\nConfirmedItemsEvo}{24}
\newcommand{\nConfirmedItemsVerified}{3}
\newcommand{\nCounterfactualBrokenHidden}{126}
\newcommand{\nCounterfactualTestedHidden}{128}
\newcommand{\nDeepSweFlagged}{7}
\newcommand{\nDeepSweTasks}{113}
\newcommand{\nEvoFlagged}{17}
\newcommand{\nEvoTasks}{48}
\newcommand{\nGFiveRequired}{5{,}339}
\newcommand{\nGradedMilestones}{98}
\newcommand{\nGradedMilestonesDubbo}{12}
\newcommand{\nGradedMilestonesElementWeb}{18}
\newcommand{\nGradedMilestonesGoZero}{23}
\newcommand{\nGradedMilestonesNavidrome}{9}
\newcommand{\nHoldoutAnnotatorsAgree}{92}
\newcommand{\nKappaCrossPairs}{553}
\newcommand{\nKappaPairsSame}{8}
\newcommand{\nMilestonesHard}{25}
\newcommand{\nMilestonesHardBeforePathRule}{34}
\newcommand{\nMilestonesHardDubbo}{1}
\newcommand{\nMilestonesHardElementWeb}{7}
\newcommand{\nMilestonesHardGoZero}{15}
\newcommand{\nMilestonesHardNavidrome}{2}
\newcommand{\nModelFamilies}{11}
\newcommand{\nPathHardDeterminate}{398}
\newcommand{\nPathHardWritten}{364}
\newcommand{\nPrecisionHoldoutK}{58}
\newcommand{\nPrecisionHoldoutN}{65}
\newcommand{\nRecallHoldoutK}{13}
\newcommand{\nRecallHoldoutN}{14}
\newcommand{\nRegObs}{3{,}379}
\newcommand{\nRegPairsPossible}{3{,}465}
\newcommand{\nRegPairsUnknown}{86}
\newcommand{\nRegRuns}{35}
\newcommand{\nRegUnitsBlocked}{72}
\newcommand{\nRegUnitsControl}{27}
\newcommand{\nReleaseFlaggedTasksSweMilestone}{31}
\newcommand{\nReviewedItemsChain}{40}
\newcommand{\nReviewedItemsDeepSwe}{23}
\newcommand{\nReviewedItemsEvo}{40}
\newcommand{\nReviewedItemsVerified}{4}
\newcommand{\nValidationCasesHoldout}{94}
\newcommand{\nValidationCasesSeedSet}{111}
\newcommand{\nVerifiedFlagged}{3}
\newcommand{\nVerifiedInstances}{500}
\newcommand{\nWaivedEntriesConservative}{3{,}321}
\newcommand{\nWaivedEntriesFull}{3{,}964}
\newcommand{\nullMedianTauResolve}{0.980}
\newcommand{\nullMedianTauScore}{0.990}
\newcommand{\nullQuantileTauResolve}{9.5}
\newcommand{\nullQuantileTauScore}{24.9}
\newcommand{\nullStratifiedQuantileResolveThirtyFive}{52.8}
\newcommand{\nullStratifiedQuantileScoreThirtyFive}{13.3}
\newcommand{\obsidianIdDiffersFailed}{174}
\newcommand{\pctAtRisk}{0.6}
\newcommand{\pctAtRiskGoZero}{7.1}
\newcommand{\pctChainItemPrecision}{70.0}
\newcommand{\pctEvoItemPrecision}{60.0}
\newcommand{\pctFailFlagged}{80.6}
\newcommand{\pctFailGroupI}{80.6}
\newcommand{\pctFailGroupIi}{52.2}
\newcommand{\pctFailGroupIii}{48.4}
\newcommand{\pctFailUnflagged}{15.1}
\newcommand{\pctLogJoint}{13.6}
\newcommand{\pctLogSole}{27.1}
\newcommand{\pctPathHardWritten}{91.5}
\newcommand{\pctPrecisionHoldout}{89.2}
\newcommand{\pctPrecisionHoldoutHigh}{94.7}
\newcommand{\pctPrecisionHoldoutLow}{79.4}
\newcommand{\pctRecallHoldout}{92.9}
\newcommand{\pctRecallHoldoutHigh}{98.7}
\newcommand{\pctRecallHoldoutLow}{68.5}
\newcommand{\pctRegRateDiff}{31.9}
\newcommand{\pctRegRateDiffHigh}{43.7}
\newcommand{\pctRegRateDiffLow}{18.3}
\newcommand{\peakRssMib}{1{,}333}
\newcommand{\regOddsRatio}{4.64}
\newcommand{\regOddsRatioBootHigh}{8.56}
\newcommand{\regOddsRatioBootLow}{2.62}
\newcommand{\regOddsRatioHigh}{7.37}
\newcommand{\regOddsRatioLow}{2.92}
\newcommand{\regOddsRatioNoLeak}{4.75}
\newcommand{\regOddsRatioNoLeakHigh}{7.54}
\newcommand{\regOddsRatioNoLeakLow}{2.99}
\newcommand{\tauResolveFullThirtyFive}{0.960}
\newcommand{\tauResolveFullTwentyEight}{0.942}
\newcommand{\tauResolveNoLeak}{0.960}
\newcommand{\tauScoreFullThirtyFive}{0.980}
\newcommand{\tauScoreFullTwentyEight}{0.984}
\newcommand{\tauScoreNoLeak}{0.979}

\title{What the Specification Did Not Say:\\Specification--Oracle Gaps in Test-Graded Coding-Agent Benchmarks}

\author{Fei Ding\thanks{Correspondence: \href{mailto:dignfei@gmail.com}{\texttt{dignfei@gmail.com}}.} \\
  Alibaba Group}

\begin{document}
\maketitle
\begin{abstract}
Coding-agent benchmarks often describe a task in natural language but grade it with tests taken from the upstream project. Passing those tests can require information the description never states---a constant's name, a module path, an exact message. We call this a \emph{specification--oracle gap}: where it exists, a benchmark also measures whether an agent reproduces upstream implementation details. We formalise the gap, including the inconsistencies that chains of dependent tasks add, and build a static audit for five languages that classifies every name-, path- and literal-level dependency of every scored test (held-out precision \pctPrecisionHoldout\%, recall \pctRecallHoldout\%). On SWE-Milestone (\nGradedMilestones{} graded milestones, 35 published runs), \nMilestonesHard{} milestones contain a test that depends on an unstated, non-inferable name or path. Build units with such a dependency fail in \pctFailGroupI\% of runs, against \pctFailGroupIii\% for comparable units without one, and withholding a single hidden name from the reference stops its unit's tests from compiling in \nCounterfactualBrokenHidden{} of \nCounterfactualTestedHidden{} cases. The leaderboard's ranking is robust to removing the affected tests, but they shift scores within repositories and favour agents that reproduce upstream names. Across four further benchmarks, prevalence appears to follow construction: a third or more of the tasks are affected where specifications are release notes or LLM summaries, almost none where tasks are screened or written for their tests; how much a gap costs depends on how the harness grades. We release the audit, corrected filter lists and specification addenda.
\end{abstract}

\section{Introduction}
\label{sec:intro}
Benchmarks for LLM coding agents have moved from isolated issues \citep{swebench} to long-horizon software evolution, in which an agent carries one code base through a sequence of dependent changes \citep{swemilestone,sweevo,sweci}. Such benchmarks are realistic, and they can be mined automatically: a pipeline groups a project's commits into milestones, writes a natural-language specification for each, and reuses the project's own tests as the oracle.

This construction joins two artefacts of different kinds. The specification describes behaviour; the oracle is code that calls helpers, references constants, imports modules at fixed paths and compares exact messages. Whatever the oracle depends on that the specification does not state, an agent may implement the specified behaviour correctly and still fail if it names, places or phrases something differently. We call this mismatch a \emph{specification--oracle gap}. Where a gap exists, the benchmark no longer measures only whether an agent implements what it was asked to implement; it also measures whether the agent reproduces upstream implementation details.

Single-task benchmarks know a version of this problem: SWE-bench Verified \citep{swebenchverified} removed tasks whose tests checked details absent from the issue. Long-horizon benchmarks raise it in a sharper form. A missing name typically fails a whole \emph{build unit}, the group of tests that is compiled or loaded together---a Go package, a Rust test target, a Maven module---rather than one test; it recurs in every later milestone whose tests use it; and the milestone chain adds gaps of its own when one milestone's reference tests depend on another milestone's code.

We ask how large these gaps are in practice, whether they actually change agents' results, and what they do to the conclusions drawn from a benchmark. Our contributions are:
\begin{itemize}
\item a definition of specification--oracle gaps for milestone benchmarks, with a taxonomy that separates under-specification (unstated, not inferable) from chain inconsistency (no agent could pass), and a model of their blast radius;
\item a static audit for Go, Python, Rust, TypeScript and Java that classifies every name-, path- and literal-level dependency of every required test, validated on a held-out sample labelled from public files (precision \pctPrecisionHoldout\%, recall \pctRecallHoldout\%) and running in minutes without model calls;
\item a study of SWE-Milestone v1.0.2 \citep{swemilestone} with its 35 published runs over 27 models: gaps affect \nMilestonesHard{} of \nGradedMilestones{} milestones; units with a gap fail \liftGroupI{} times as often as comparable units, and a counterfactual on the reference confirms the predicted blast radius; removing the affected tests leaves the leaderboard's ranking within what random removals of the same size produce, but shifts scores within the most affected repository; agents' reproduction of hidden names is a stable trait of the model, unrelated to its training cutoff;
\item a comparison with four benchmarks built in other ways (SWE-EVO, SWE-Chain, SWE-bench Verified, DeepSWE), using only public data: prevalence follows how the task text is made, and the harness decides how much a gap costs;
\item corrected filter lists and specification addenda for the benchmark, and recommendations for building test-graded benchmarks.
\end{itemize}
Our in-depth claims concern one benchmark of seven repositories; the comparison with four further benchmarks shows that the audit transfers and that prevalence depends on construction.

\section{Background: SWE-Milestone}
\label{sec:background}
SWE-Milestone \citep{swemilestone} reconstructs a stretch of a repository's release history into a directed acyclic graph of \emph{milestones}, functionally coherent groups of upstream commits. Version 1.0.2 covers seven repositories in five languages: go-zero (Go), navidrome (Go and TypeScript), nushell and ripgrep (Rust), scikit-learn (Python), element-web (TypeScript) and dubbo (Java). In the end-to-end mode used for its leaderboard, an agent starts from a \emph{trial base}---the first release of the range, plus a small patch that makes the environment build---and implements all milestones in one working tree, choosing the order among milestones whose prerequisites are done. For each milestone it receives a software requirements specification (SRS) that an LLM pipeline generates from the milestone's commits.

Grading is test-based. For each milestone the grading harness takes the upstream (\emph{gold}) state at the end of the milestone, copies its test code onto the agent's code and runs the suite. Tests are classified by their outcome on the gold states before and after the milestone, over repeated runs: \emph{fail-to-pass} and \emph{none-to-pass} tests (failing or absent before, passing after) are the new behaviour the milestone must deliver; \emph{pass-to-pass} tests must keep passing; tests that pass before but no longer run after (\emph{pass-to-none}) are not graded. Per-milestone filter lists remove tests the maintainers found invalid. Together these are the \emph{required} tests. For each milestone, recall is the share of required new tests that pass, precision is $(f+1)/(f+b+1)$ for $f$ new tests passed and $b$ pass-to-pass tests broken, and the \emph{score} is their F1; a milestone is \emph{resolved} when all required tests pass. The leaderboard averages both over the graded milestones of each repository and then over repositories; the best published configuration reaches a resolve rate of \bestResolveOriginal\% over the \nGradedMilestones{} graded milestones.

\section{Motivating Example}
\label{sec:motivation}
\input{figures/en/hidden-name-gap-figure}
Figure~\ref{fig:hidden-name-gap} illustrates how a hidden name blocks a whole build unit and recurs across milestones.
Milestone M019 of go-zero adds Server-Sent Events support. Its SRS states the behaviour precisely: ``detect SSE requests by checking for the \texttt{Accept: text/event-stream} header'' and bypass the timeout handler. It does not state the names the reference introduces for this behaviour, \texttt{headerAccept} and \texttt{valueSSE}, but the reference tests of the \texttt{rest/handler} package use both. Because Go compiles a package's tests together, an agent that implements the behaviour correctly but names the constant differently loses every required test of the package: 84 in M019, and between 88 and 108 in each of five other milestones that inherit the tests. The published runs show what is at stake. The runs' source snapshots of M019 show what each wrote: 32 of 34 runs\footnote{Counts are over 34 of the 35 runs. One run, whose name in the release marks it as having had access to leaked information, is left out here and in Section~\ref{sec:memorization}.} implement the bypass the specification asks for. Of the nine that also name the value \texttt{valueSSE}, four pass all 84 required tests (the other five lack a further name). The other 23 name it differently (\texttt{valueEventStream} in 17, \texttt{valueTextEventStream} in 2, \texttt{valueSse} in 1) or write the string inline, and pass none; for 13 of them the evaluation log reports no undefined name in the package other than \texttt{valueSSE} and \texttt{headerAccept}. Differences of this kind are invisible in an aggregate score yet decide comparisons between individual runs.

\section{Specification--Oracle Gaps}
\label{sec:approach}
\subsection{Definition}
A test-graded coding task $t$ gives the agent a specification $S_t$ in natural language and a code base $B_t$ to start from, and grades the result with an oracle $O_t$, a set of tests taken from a reference solution. Each test $o \in O_t$ \emph{depends} on a set $D(o)$ of artefacts that the solution must provide for $o$ to pass: identifiers it references (functions, types, constants, methods, fields), module or file paths it imports or mocks, and literals it compares against output (messages, formatted text). An artefact is \emph{fixed by the task} if it is defined in $B_t$ or named in $S_t$; we write $F_t$ for this set and add the artefacts that conventions already present in $B_t$ determine (\emph{inferable} names, below). Any other artefact is a free choice: an implementation that chooses it differently is equally correct. The \emph{specification--oracle gap} of a test is
\[
G(o) = D(o) \setminus F_t .
\]
A test with $G(o) \neq \emptyset$ can fail an agent that implements the specified behaviour correctly but names, places or phrases an artefact differently. The definition is not specific to long horizons; it applies to any benchmark that pairs natural-language tasks with reference tests. Behaviour that the specification leaves open lies outside $D(o)$ and our audit.

\textbf{Milestone chains.} In a milestone benchmark the agent works through milestones $m_1, m_2, \dots$ in one tree. What it can know for milestone $m$ is the \emph{trial base} it started from---not $m$'s reference start state, which already contains other milestones' reference code that the agent reimplemented in its own way---and the specifications of $m$ and of its ancestors in the dependency graph. A specification \emph{names} an artefact if it states it verbatim or as its words in order (\texttt{SetSSESlowThreshold} $\sim$ ``set SSE slow threshold''); a module path only if it names the file.

\textbf{Blast radius.} A gap rarely costs one test. If $G(o)$ contains a name or a path, every test in $o$'s build unit fails with it: the language compiles or collects the unit as a whole (Appendix~\ref{app:languages}). If $G(o)$ contains only literals, $o$ alone fails. The tests at risk from a gap are therefore those of the units it blocks, and in build systems that stop at the first failing unit, possibly a whole milestone.

\subsection{Where gaps come from}
For every artefact in $D(o)$ outside the trial base, we determine where it comes from by diffing the reference states of all milestones (Table~\ref{tab:taxonomy}). Classes A1 and A2 are under-specification: the agent could only reproduce the dependency by guessing or from memory. Class B is inconsistency along the milestone chain. B3 is the starkest: the reference test references a name that the gold end state of the same milestone does not define, so even the reference implementation cannot pass it; the benchmark has silently dropped such tests from the required set as pass-to-none. A further, order-dependent form, an \emph{order conflict}, arises because milestone golds change shared APIs incompatibly (go-zero M003 makes \texttt{RollingWindow} generic; M024 removes \texttt{setupInterceptors}): whichever milestone an agent implements first breaks the reference tests of milestones it implements later.

\begin{table}[t]
\centering\small
\begin{tabular}{@{}lp{0.68\columnwidth}@{}}
\toprule
class & the dependency is \\
\midrule
visible & defined in the trial base \\
specified & introduced by this milestone or an ancestor and named in its SRS \\
A1 & introduced by this milestone's gold change, not named in its SRS \\
A2 & introduced by an ancestor's gold change, not named in that SRS or this one \\
B1 & introduced by a milestone that is not an ancestor \\
B2 & introduced by upstream commits that belong to no milestone \\
B3 & not defined by the gold end state of the milestone itself \\
\bottomrule
\end{tabular}
\caption{Classes of test dependencies.}
\label{tab:taxonomy}
\end{table}

Not every hidden name is a guess in the dark. A dependency is \emph{inferable}, and not counted as hard, if (i) it combines a name of the trial base with an affix the code base already uses systematically (\texttt{Allow} $\rightarrow$ \texttt{AllowCtx}, with 227 such pairs in go-zero); (ii) it is a new module or file whose name the specification gives, placed in an existing directory or beside existing ones, as front-end code bases do by convention; or (iii) it is a method whose signature matches an interface of the trial base or one the specification names. Conversely, a dependency can hide in a test's identifier rather than its code: if part of a scored test's name is computed from a string that only the reference defines, every test carrying it depends on that string. Rules (ii), (iii) and the identifier rule were fixed on development data before the held-out set was drawn: (ii) encodes a front-end convention, the others came from the DeepSWE review.

\subsection{Implementation}
\input{figures/en/audit-overview-figure}
Figure~\ref{fig:audit-overview} summarises the static audit.
One extractor per language reports the names, module paths and literals that each state defines and each test references; a language-independent core diffs the gold states, classifies each dependency and maps required tests to build units (details in Appendix~\ref{app:languages}). The audit runs in under half an hour without model calls (RQ4).

\section{Experimental Setup}
\label{sec:design}
We study five questions. \textbf{RQ1} (prevalence): how many required tests and milestones depend on unstated, non-inferable information? \textbf{RQ2} (realisation): do these dependencies cause failures? \textbf{RQ3} (impact): how do scores and rankings change when the affected tests are removed? \textbf{RQ4} (precision and cost): how accurate and how costly is the audit? \textbf{RQ5} (other benchmarks): does the gap appear in benchmarks built differently?

\textbf{Subjects and runs.} All seven repositories of SWE-Milestone v1.0.2 (\nGradedMilestones{} graded milestones, 580,039 required tests) and its 35 published end-to-end runs (Claude Code, Codex, Gemini CLI and OpenHands \citep{openhands} with 27 models), whose per-test results are public; replayed through the harness's own scoring code, our scores match the 28 leaderboard runs to within 0.20 points. One run whose release name marks it as having had access to leaked information is kept in aggregate results and left out wherever we ask which names runs reproduced. Leaving it out everywhere changes no conclusion (it is not among the 28 leaderboard runs): the lift and odds ratio of RQ2 become \liftSameNoLeak{} and \regOddsRatioNoLeak, and Kendall's $\tau$ over all runs (Table~\ref{tab:rescore_summary}) \tauScoreNoLeak{} by score and \tauResolveNoLeak{} by resolve rate.

\textbf{Realisation.} A unit is a (milestone, build unit) pair; it \emph{fails} in a run when one of its required tests fails or is missing. Since units with hidden dependencies are units whose tests exercise new code, we compare them with units that also carry fail-to-pass or none-to-pass tests but no new dependency, within the same milestones, with 95\% intervals from a bootstrap over milestones; we also read each failure's log for what it reports as missing, and, for Go, rebuild the reference with one hidden name renamed to see whether the predicted unit breaks.

\textbf{Re-scoring} removes (a) the tests in units blocked by a hard hidden dependency and (b) tests whose reference lacks a name that no environment patch supplies (\emph{conservative}), and in the \emph{full} variant also (c) tests that no run passes and whose cause is explained (Appendix~\ref{app:allfail}); we recompute the metrics with the harness, compare rankings with Kendall's $\tau$-b, bootstrap over milestones, and compare against removing as many tests from randomly chosen units of matched size.

\textbf{Other benchmarks.} SWE-EVO \citep{sweevo} and SWE-Chain \citep{swechain} (release-level tasks), SWE-bench Verified \citep{swebenchverified} and DeepSWE~1.1 \citep{deepswe}. The mapping is direct: $S$ is the task text the agent sees, $B$ the repository at the base commit, the gold change the reference patch, and the oracle the scored tests. For realisation we use only public results: SWE-Chain's per-step results of nine systems, the per-instance results of the 78 SWE-bench Verified submissions in the official experiments repository that resolve at least 60\%, and DeepSWE's released trials.

\section{Results}
\label{sec:results}
\subsection{RQ1: Prevalence}
\begin{table}[t]
\centering
{\scriptsize\setlength{\tabcolsep}{2pt}\begin{tabular}{lrrrrr}
\toprule
Repository & M & Required & Hidden & Tests at risk \% & Hard M \\
\midrule
go-zero (Go) & 23 & 43{,}253 & 132 & 7.1 & 15/23 \\
navidrome (Go+TS) & 9 & 13{,}642 & 2 & 0.3 & 2/9 \\
element-web (TS) & 18 & 94{,}114 & 7 & 0.1 & 7/18 \\
nushell (Rust) & 13 & 67{,}504 & 0 & 0.0 & 0/13 \\
dubbo (Java) & 12 & 82{,}863 & 1 & 0.0 & 1/12 \\
ripgrep (Rust) & 11 & 11{,}964 & 0 & 0.0 & 0/11 \\
scikit-learn (Py) & 12 & 266{,}699 & 0 & 0.0 & 0/12 \\
\midrule
All & 98 & 580{,}039 & 142 & 0.6 & 25/98 \\
\bottomrule
\end{tabular}}
\caption{Hidden dependencies over the graded milestones (M), language in parentheses. Hidden: hard hidden names (unnamed by the SRS, absent from the trial base, not inferable), summed per milestone; Tests at risk: share of the required tests in units they block or with a hidden literal or test-level name; Hard M: milestones with at least one hard hidden dependency.}
\label{tab:prevalence}
\end{table}
Table~\ref{tab:prevalence} summarises the audit. \nMilestonesHard{} of the \nGradedMilestones{} graded milestones contain a required test with a hard hidden dependency; because a milestone is resolved only if all its required tests pass, an agent that does not happen to reproduce the hidden items cannot resolve them. The tests at risk are few overall (\pctAtRisk\% of the required tests) but concentrated: in go-zero, whose tests exercise package internals, \nMilestonesHardGoZero{} of \nGradedMilestonesGoZero{} milestones are affected and \pctAtRiskGoZero\% of the required tests sit in blocked units; element-web (\nMilestonesHardElementWeb{} of \nGradedMilestonesElementWeb{}), navidrome (\nMilestonesHardNavidrome{} of \nGradedMilestonesNavidrome{}) and dubbo (\nMilestonesHardDubbo{} of \nGradedMilestonesDubbo{}) follow, while the repositories whose tests drive public interfaces (scikit-learn, ripgrep) and nushell have none. Most hard items are A2: names an ancestor milestone introduced without its specification stating them, which later milestones' tests rely on. The rules of Section~\ref{sec:approach} that make convention-placed paths inferable and read all ancestors' specifications reduce the count from \nMilestonesHardBeforePathRule{} milestones to \nMilestonesHard{}.

Blast radius explains why a few names matter. In go-zero, two constants, \texttt{headerAccept} and \texttt{valueSSE}, block the \texttt{rest/handler} package in six milestones, 84 to 108 required tests each (Section~\ref{sec:motivation}).

Chain inconsistency is a second, independent source. No published run built nushell's milestone G05\_be6e868: its reference tests initialise a struct without a field, \texttt{custom\_completion}, that milestone G01\_48bca0a adds, so the 19 runs that had implemented G01\_48bca0a first could not compile them, and the rest stopped at other build errors in the same tests. The reference tests of go-zero's M017 use names that M017's own gold state does not define (e.g.\ \texttt{maxFormParamCount}), so even the reference cannot compile them; the benchmark silently recorded such tests as pass-to-none and dropped them from the required set (Appendix~\ref{app:cases}).

\subsection{RQ2: Realisation}
Hidden dependencies are associated with failures, though less strongly than a naive comparison suggests. Across the 35 runs, units blocked by a hard hidden dependency fail in \pctFailFlagged\% of (unit, run) pairs and all other units in \pctFailUnflagged\%, but most unflagged units only hold pass-to-pass tests, which an agent passes by breaking nothing. Within the same milestones, units that carry new tests without any new dependency fail in \pctFailGroupIii\%, and units whose new names the specification states in \pctFailGroupIi\%; blocked units fail \liftGroupI{} times as often as the former (95\% CI \liftGroupILow--\liftGroupIHigh), while stated names make no difference (\liftGroupIi, CI \liftGroupIiLow--\liftGroupIiHigh). The contrast survives the dependence between runs: a logistic regression with unit size and repository as covariates gives an odds ratio of \regOddsRatio{} (95\% CI \regOddsRatioLow--\regOddsRatioHigh, errors clustered by milestone and model family, Appendix~\ref{app:regression}; odds ratios exceed rate ratios when nearly half of the comparison units fail), and blocked units fail \matchedRatio{} times as often as new-test units of the same repository and size (CI \matchedRatioLow--\matchedRatioHigh). The logs agree: in \pctLogSole\% of the blocked units' failures, the hidden items are the only names the run's log reports as missing, and in another \pctLogJoint\% they are among them; most of the rest report no name at all.

The counterfactual isolates the mechanism without agents. Renaming one hidden identifier in the reference solution's non-test code, a counterfactual for an agent that implements the behaviour under another name, stops the unit's tests from compiling in \nCounterfactualBrokenHidden{} of \nCounterfactualTestedHidden{} Go cases, and the tests that then fail are 99.5\% of those the audit predicts.

\subsection{RQ3: Impact on scores and rankings}
\begin{table}[t]
\centering
{\scriptsize\setlength{\tabcolsep}{2pt}\begin{tabular}{lrrrrrrrr}
\toprule
Var. & Runs & $\bar\Delta$S & $\bar\Delta$R & $\Delta$R$_{\max}$ & $\tau_S$ & $\tau_R$ & Mv. & Max \\
\midrule
Full & 35 & +0.83 & +0.51 & +2.78 & 0.980 & 0.960 & 11 & 2 \\
 & 28 & +0.84 & +0.47 & +2.78 & 0.984 & 0.942 & 5 & 2 \\
\midrule
Cons. & 35 & +0.37 & +0.33 & +2.04 & 0.987 & 0.979 & 8 & 1 \\
 & 28 & +0.39 & +0.30 & +2.04 & 0.995 & 0.972 & 2 & 1 \\
\bottomrule
\end{tabular}}
\caption{Re-scoring without hidden-dependency tests. S: Score (macro F1), R: Resolve (\%); mean changes in points against the same runs scored with the v1.0.2 lists, $\Delta$R$_{\max}$ the largest; $\tau_S$, $\tau_R$: Kendall $\tau_b$ with the original ranking, by Score and by Resolve; Mv.: runs whose Score rank changes; Max: largest rank change. Full waives (a) hidden-dependency units, (b) tests whose reference lacks a name (none in v1.0.2) and (c) explained all-fail tests; Cons. (conservative) only (a) and (b).}
\label{tab:rescore_summary}
\end{table}
Removing the affected tests moves absolute numbers a little and rankings hardly at all (Table~\ref{tab:rescore_summary}). In the full variant the 28 leaderboard runs gain \deltaScoreMeanFullTwentyEight{} score points on average (\deltaScoreMinFullTwentyEight{} to \deltaScoreMaxFullTwentyEight) and \deltaResolveMeanFullTwentyEight{} resolve points (up to \deltaResolveMaxFullTwentyEight). Kendall's $\tau$-b with the published ranking is \tauScoreFullTwentyEight{} by score and \tauResolveFullTwentyEight{} by resolve rate. Over all 35 runs (\tauScoreFullThirtyFive{} and \tauResolveFullThirtyFive), these changes are within what chance produces: removing as many tests from randomly chosen units of matched size gives a median $\tau$ of \nullMedianTauScore{} (score) and \nullMedianTauResolve{} (resolve), and a $\tau$ as low as the real one occurs in \nullQuantileTauScore\% (score) and \nullQuantileTauResolve\% (resolve) of the random draws. Matching the random units also on repository, milestone, unit type and baseline failure rate gives \nullStratifiedQuantileScoreThirtyFive\% and \nullStratifiedQuantileResolveThirtyFive\%. The leaderboard's own resolution is the tighter limit: bootstrapping over milestones, all \nAdjacentPairs{} pairs of neighbouring runs among the 35 have overlapping 95\% rank intervals, so the benchmark does not separate neighbours with or without the correction.

The gap matters where it concentrates. On go-zero, the mean score of the 35 runs rises from \goZeroScoreMeanOfficial{} to \goZeroScoreMeanFull{} and the mean resolve rate from \goZeroResolveMeanOfficial\% to \goZeroResolveMeanFull\%, which studies that report per-repository results or ablate on a single repository inherit. The effect may also be partly a property of the model: two runs of the same model in different configurations agree on which hidden names they write with median Cohen's $\kappa$ = \kappaSameModelMedian{} (\nKappaPairsSame{} pairs), against \kappaCrossModelMedian{} for runs of different models (\nKappaCrossPairs{} pairs). With so few same-model pairs the difference is not significant, so this is exploratory (Appendix~\ref{app:agreement}).

\subsection{RQ4: Precision and cost}
We validate on (milestone, build unit) cases sampled from public data in three strata per repository---units the audit flags, unflagged units that fail as a whole in some published run (where misses would show) and unflagged units that mostly pass---and labelled from the specifications, the reference change and the tests alone. A first set of \nValidationCasesSeedSet{} cases exposed implementation errors (ancestor specifications not read, literals not normalised); after fixing them we drew a second, disjoint set of \nValidationCasesHoldout{} cases and then all 48 remaining flagged units with 55 unflagged ones (103 cases), which a hired software expert labelled blind before comparing. Over the 65 flagged units of these two held-out rounds the audit has precision \nPrecisionHoldoutK/\nPrecisionHoldoutN{} (\pctPrecisionHoldout\%; 95\% Wilson interval \pctPrecisionHoldoutLow--\pctPrecisionHoldoutHigh\%); on the second set its recall is \nRecallHoldoutK/\nRecallHoldoutN{} (\pctRecallHoldout\%; \pctRecallHoldoutLow--\pctRecallHoldoutHigh\%). One of the authors labelled the same cases independently (Cohen's $\kappa$ = \kappaHoldoutAnnotators{} and 0.96; \nHoldoutAnnotatorsAgree{} of \nValidationCasesHoldout{} and 101 of 103 cases agree); against the author's labels, precision is 56/65. Its seven false positives are dependencies the specification implies without stating them---modules named or placed by convention, packages it says are moved, a method fixed by a protocol it names---and one identifier hidden behind Markdown escapes; we added no rule for them after seeing the sets,. Against agent behaviour, precision is lower by design: a flag means that the specification withholds a dependency, not that every agent fails (Section~\ref{sec:memorization}). A complete audit of the seven repositories from a cold cache takes \minutesColdAudit{} minutes on a 16-core workstation, with at most \peakRssMib{}\,MiB of memory and no model calls; repeated runs give identical results.

\subsection{RQ5: Other benchmarks}
\label{sec:other}
\begin{table}[t]
\centering\footnotesize\setlength{\tabcolsep}{3pt}
\begin{tabular}{@{}llrrr@{}}
\toprule
Benchmark & Spec. & Flagged & Prec. & Lift \\
\midrule
SWE-Milestone & LLM & \nMilestonesHard/\nGradedMilestones & \nPrecisionHoldoutK/\nPrecisionHoldoutN{} (\pctPrecisionHoldout\%) & \liftGroupI \\
SWE-EVO & notes & \nEvoFlagged/\nEvoTasks & \nConfirmedItemsEvo/\nReviewedItemsEvo{} (\pctEvoItemPrecision\%) & -- \\
SWE-Chain & LLM & \nChainFlagged/\nChainSteps & \nConfirmedItemsChain/\nReviewedItemsChain{} (\pctChainItemPrecision\%) & \chainLift \\
SWE-bench Verif. & issue & \nVerifiedFlagged/\nVerifiedInstances & \nConfirmedItemsVerified/\nReviewedItemsVerified{} & 3.4 \\
DeepSWE & written & \nDeepSweFlagged/\nDeepSweTasks & \nConfirmedItemsDeepSwe/\nReviewedItemsDeepSwe{} & -- \\
\bottomrule
\end{tabular}
\caption{The audit on five benchmarks. Spec.: how the task text is made (LLM summary, release notes, screened issue, written for it). Flagged: tasks with a hidden name, path or literal (SWE-Milestone: hard names or paths, \nReleaseFlaggedTasksSweMilestone{} with literals; SWE-Chain: A1 only); Prec.: reviewed flags, per held-out unit for SWE-Milestone (RQ4), per item elsewhere, not comparable across rows (Verified: 3 of 4 items, in 2 of 3 instances); Lift: failure-rate ratio---SWE-Milestone: blocked build units against comparable units (RQ2); SWE-Chain: flagged against unflagged steps; Verified: name-flagged against unflagged instances; no lift for SWE-EVO (no per-task results) or DeepSWE (one confirmed).}
\label{tab:benchmarks}
\end{table}
Table~\ref{tab:benchmarks} applies the unchanged audit to four more benchmarks (Python only except DeepSWE). The comparison is exploratory: review units and sizes differ (Appendix~\ref{app:reviews}). Prevalence follows construction. Where the specification is a release note (SWE-EVO) or an LLM summary told to leave out ``exact messages'' and ``internal identifiers'' (SWE-Chain), a third or more of the tasks depend on something it does not say---a module split the note never mentions, an exception message, a warning class that only the reference names; SWE-Chain's chains also show A2 dependencies, names an earlier step introduced and no later specification states. Where issues were screened by annotators (SWE-bench Verified) or instructions were written for the tests (DeepSWE), hidden names are almost absent: of DeepSWE's flags, only one survives review; the others are names its instructions give in prose and literals the tests pass as input.

\textbf{Rare, but decisive.} The two confirmed Verified instances import a name the issue never mentions (\texttt{get\_annotation} and \texttt{infer\_node}; \texttt{IS\_PYPY}), each failing all 10 or 21 fail-to-pass tests at collection; none of 78 strong public submissions resolves either. In DeepSWE, the one gap hides in test \emph{identifiers}: a test helper names every case after the rule's display name, ``Auto Table of Contents'', which only the reference defines while the instruction gives the alias \texttt{auto-toc}. The task has the lowest pass rate of all \nDeepSweTasks{} (2.5\% of 280 public trials, 28 models); in \obsidianIdDiffersFailed{} of the 248 trials with a test report the cases ran under another group name---163 times ``Auto TOC''---and all 41 scored identifiers counted as failed. Persistent failures in Verified mostly have other causes: of 40 unflagged instances that at most 20\% of the submissions resolve, our review attributes 28 to behaviour the issue leaves open, four to test defects, three to items outside the audit's rules and five to the systems.

\textbf{The harness decides whether a gap bites.} SWE-Chain flags a larger share of its tasks than SWE-Milestone, yet flagged steps fail only slightly more often than others (lift \chainLift, 95\% CI \chainLiftLow--\chainLiftHigh); its harness isolates failed imports and reports them back to the agent (Appendix~\ref{app:harness}).

\subsection{Reproduced hidden names}
\label{sec:memorization}
A flag means that the specification withholds a dependency, not that every agent misses it. Runs reproduce 45\% of the hard dependencies for which their evaluation gives evidence and 79\% of the inferable ones, and we find no sign that reproduction depends on whether a model's training cutoff postdates the upstream code (Appendix~\ref{app:memorization}).

\section{A Corrected Benchmark and Recommendations}
\label{sec:verified}
\textbf{Corrected filter lists.} For every graded milestone we extend the v1.0.2 filter list with entries that waive (a) the required tests in units blocked by a hard hidden dependency and (b) tests whose reference lacks a name that no environment patch supplies; the full variant also waives (c) tests that no published run passes and whose cause the audit explains. The lists pass the harness's own validation and can replace the benchmark's; they hold \nWaivedEntriesConservative{} (conservative) or \nWaivedEntriesFull{} (full) entries, each with its reason.

\textbf{Specification addenda.} Waiving tests discards evidence. The alternative is to tell the agent what the tests need. For each of the \nMilestonesHard{} affected milestones we generate an addendum that states the interface conventions the reference tests rely on---\nAddendumConventions{} in total---without describing their implementation, e.g.\ ``the package \texttt{rest/handler} must declare \texttt{valueSSE} (relied on by 84 required tests)''. Addenda keep the tests and remove the guessing; evaluating them requires new agent runs.

\textbf{Why the generator leaves names out.} SWE-Milestone's published generation prompts tell the model not to over-specify implementation details, and its test-aligned revision pass may omit details the agent can ``figure \ldots{} out from the code diff''---which the agent never sees (Appendix~\ref{app:generator}).

\textbf{Recommendations.} (1) Run a dependency audit when generating specifications, and either state the names, paths and messages the tests need or exclude tests that reach into internals. (2) Replay the reference solutions along the chain in several topological orders and repair tests that fail for any order. (3) Independently of the tests, isolate build units during grading so that one unit's compile failure cannot erase a milestone, and report unresolved names back to the agent, as SWE-Chain's harness does. (4) Publish per-test results, as SWE-Milestone does; without them neither our realisation analysis nor the re-scoring would be possible.

\section{Related Work}
\label{sec:related}
\textbf{Validity of coding benchmarks.} SWE-bench \citep{swebench} established test-based grading of agent patches. SWE-bench Verified \citep{swebenchverified} had annotators exclude tasks with under-specified issues or overly specific tests; a later audit by its authors found that most of a subset of hard instances still had tests that reject correct solutions \citep{openai2026verified}, and our review of persistently failed instances points the same way. SWE-Bench+ \citep{swebenchplus} reported solution leakage and weak tests; EvalPlus \citep{evalplus}, UTBoost \citep{utboost} and \citet{solvedissues} showed that passing tests can still accept wrong code; PAIChecker \citep{paichecker} detects misalignment between issues and their pull requests; SWE-Bench Pro Verified \citep{sweproverified} and SciCode-Verified \citep{scicodeverified} re-annotated benchmarks and found that defects understated model ability; \citet{agenticbest} propose checklists for agentic benchmarks. These works audit single tasks, mostly by manual annotation, and mostly the opposite failure: tests that accept too much. We study tests that demand too much, measure it statically across five languages, and follow its effect through build units and milestone chains.

\textbf{Telling the agent what the tests need.} Several benchmarks state interfaces in the task: SWE-Bench Pro \citep{swebenchpro} adds requirements and interface descriptions to each task, FEA-Bench \citep{feabench} gives the signatures of new components, and Commit0 \citep{commit0} provides a library's specification documents. DeepSWE \citep{deepswe} writes instructions and verifiers for each task. Our addenda bring this practice to a mined long-horizon benchmark, and the audit shows where it is needed.

\textbf{Long-horizon and multilingual benchmarks.} SWE-Milestone \citep{swemilestone}, SWE-EVO \citep{sweevo} and SWE-CI \citep{sweci} evaluate agents over sequences of changes; Multi-SWE-bench \citep{multiswe} and SWE-PolyBench \citep{polybench} extend issue resolution beyond Python; SWE-bench-Live \citep{swebenchlive} keeps adding fresh tasks against contamination.

\textbf{Benchmarks that measure the wrong thing.} In NLP, annotation artefacts let models pass natural language inference without inference \citep{gururangan}, and benchmark design has been called a validity problem in its own right \citep{bowmandahl}; contamination \citep{sainz} and memorisation of SWE-bench \citep{illusion} raise the question of recall versus skill that our reproduction analysis touches. A specification--oracle gap is the generative analogue of an annotation artefact: the grader rewards information that the input does not contain.

\section{Conclusion}
Generated specifications and verbatim tests make benchmarks cheap to build and realistic to run, but they let tests depend on information the task does not state. In SWE-Milestone a quarter of the graded milestones contain such a dependency; the build units that carry one fail markedly more often than comparable units, and renaming a single hidden identifier in the reference breaks its unit just as the audit predicts. The leaderboard's ranking survives the correction---and is, in any case, coarser than it looks---but per-repository scores shift. Across five benchmarks the gap follows how the task text is made. A dependency audit is cheap---minutes, no model calls---and we release ours with corrected filter lists and specification addenda.

\section*{Limitations}
A hidden dependency is a risk, not a certain failure: agents reproduce some unspecified names, from conventions or from memory of the upstream project, so we report realisation separately and mark names derivable from base conventions as inferable. Realisation is observational: we compare blocked units with units that also test new code, within the same milestones, but the groups can still differ in difficulty; the counterfactual on the reference covers Go only. Three of our rules (convention-placed paths, interface methods, identifier-level dependencies) and the fixes of RQ4 were set on development data and frozen before the held-out set was drawn; we report the counts before and after (RQ1). DeepSWE, whose review prompted two rules, is not an out-of-sample test. Our SRS matching is lexical; a specification may describe a name in other words (over-reporting) or mention an identifier without meaning it as an interface (under-reporting). The extractors are name-level and do not type-check: they miss a new method whose name already exists on another type in the same unit, and they can flag interface methods required by external libraries. Behaviour that tests require without new names (an escaping rule, a weighting scheme) is covered only by a text heuristic that we report apart; on SWE-bench Verified such behavioural gaps explain most persistent failures we reviewed, so the audit's counts are a lower bound on how often tests demand what the task does not say. Dependencies outside the rules---keyword arguments, attribute access on objects in Python and TypeScript, strings compared through helpers---are reported only as name-level upper bounds, and test identifiers computed at run time can only be matched where the auditor reads the helper that builds them. Flagged items and persistent failures were reviewed against the task text, the tests and the reference (with LLM-drafted first-pass verdicts checked by the authors); such judgements can differ, and we release every verdict with its quotations. We study one benchmark in depth and four more by comparison; prevalence will vary with language, test style and construction, as it does across our repositories and benchmarks. The memorisation analysis is correlational. Removing tests changes what a benchmark measures; we therefore also provide specification addenda, whose effect we have not yet measured with new agent runs.

\section*{Ethics Statement}
We analyse publicly released benchmark data and agent runs; no human-subjects research is involved and we run no coding agents. The held-out validation labels come from one author and a hired software expert with over ten years of full-stack development and open-source maintenance experience, recruited through personal contacts, paid US\$80 per hour, well above typical annotation rates, and informed that the labels would be released; participation was voluntary. All artefacts we use are public: SWE-Milestone, its harness and its run logs (MIT), DeepSWE's tasks and published trials (Apache-2.0, each task's upstream project under its own permissive license), SWE-bench Verified (distributed with the MIT-licensed SWE-bench), the per-instance resolution records of the SWE-bench experiments repository, which states no license and from which we use only whether each instance was resolved, SWE-EVO (MIT; its dataset card also names Apache-2.0) and SWE-Chain (Apache-2.0). We use them for evaluation research, their intended use, and redistribute none of their contents beyond identifiers and short quotations. The audit could be used to tailor agents to a benchmark's hidden names; we release it to improve benchmarks, and the corrected filter lists remove the advantage such tailoring would give. \textbf{Use of AI assistants.} We used LLM-based coding assistants to help implement the audit and the analysis scripts, to draft first-pass verdicts for the reviews of flagged and persistently failed cases, and to edit the text. The definitions and design decisions are the authors'; every review verdict and every reported number was checked by the authors against the released data and code.

\appendix
\section{Units and Extractors per Language}
\label{app:languages}
The audit has one extractor per language: Go's \texttt{go/ast} with object resolution and build constraints; Python's \texttt{ast}; tree-sitter for TypeScript, Rust and Java (Table~\ref{tab:languages}). Extractors report, per state, the names each package or module defines and, per test unit and test, the names, module paths and literals it references. A language-independent core diffs the gold states of all milestones, classifies each dependency, applies the SRS and inferability rules and maps required tests to units. Required test sets are computed with the harness's own normalisation code and match the official counts on all milestones. Gold states are exported from the benchmark's images.

\begin{table*}[t]
\centering\footnotesize
\setlength{\tabcolsep}{4pt}
\begin{tabular}{@{}lllp{0.55\textwidth}@{}}
\toprule
Language & Extractor & Unit & Notes \\
\midrule
Go & go/ast & package test binary & build constraints for linux/amd64; Ginkgo suites one unit \\
Python & ast & test module & module-level import failure fails collection; C extensions read by pattern \\
TypeScript & tree-sitter & test file & imports, \texttt{jest.mock}/\texttt{vi.mock} paths; path specified only if the SRS names the file \\
Rust & tree-sitter & cargo test target & macros and rstest cases mapped; workspace build stops at first failing target \\
Java & tree-sitter & Maven module & the multi-module build skips modules after a failed one \\
\bottomrule
\end{tabular}
\caption{Build units and extractors per language.}
\label{tab:languages}
\end{table*}

\section{Regression Details}
\label{app:regression}
The model of RQ2 is a logistic regression of whether a (unit, run) pair fails---a required test fails or is missing---on an indicator for a hard-blocked unit, the base-2 logarithm of the unit's number of required tests, and repository fixed effects. It is fitted over the \nMilestonesHard{} milestones that contain a hard-blocked unit, on the \nRegObs{} pairs of their \nRegUnitsBlocked{} blocked and \nRegUnitsControl{} comparison units (new tests, no new dependency) over \nRegRuns{} runs; \nRegPairsUnknown{} of the \nRegPairsPossible{} pairs have no readable result and are dropped. Each run is a separate observation; runs of the same model in other configurations are not merged but share a model family. Standard errors are two-way cluster-robust over milestones and \nModelFamilies{} model families (vendor plus tier, e.g.\ Claude Opus or Gemini Flash); a two-way cluster bootstrap that resamples milestones and families independently (500 draws) gives \regOddsRatioBootLow--\regOddsRatioBootHigh. Setting the blocking indicator to zero lowers the predicted failure probability of blocked pairs by \pctRegRateDiff{} points on average (\pctRegRateDiffLow--\pctRegRateDiffHigh). Without the run marked as leaked, the odds ratio is \regOddsRatioNoLeak{} (\regOddsRatioNoLeakLow--\regOddsRatioNoLeakHigh).

\section{Agreement Between Runs}
\label{app:agreement}
For each pair of runs we compute Cohen's $\kappa$ on whether each hard hidden name was written, over the names for which both runs' evaluations give evidence (Appendix~\ref{app:memorization}); the run marked as leaked is left out. Same-model pairs are runs of the same model id in two configurations (\nKappaPairsSame{} pairs, all other runs having a single configuration); cross-model pairs are runs of different models (\nKappaCrossPairs{} pairs). The medians are \kappaSameModelMedian{} (95\% interval \kappaSameModelLow--\kappaSameModelHigh) and \kappaCrossModelMedian{} (\kappaCrossModelLow--\kappaCrossModelHigh), from 500 bootstrap draws over the hard names; the interval of their difference, \kappaDiffLow{} to \kappaDiffHigh, comes from a bootstrap over models.

\section{Tests No Run Passes}
\label{app:allfail}
Of the \nAllFailTotal{} required tests that no published run passes, \nAllFailNoBuild{} are attributed to the nushell milestone no run could build (its remaining test is counted under hidden literals), \nAllFailEnvironment{} to tests no run executed because of the environment, \nAllFailAOne{} and \nAllFailATwo{} to A1 and A2 dependencies, \nAllFailLiteral{} to hidden literals and \nAllFailBehaviour{} to behaviour no specification describes; \nAllFailUnexplained{} remain unexplained.

\section{Why the Generator Leaves Names Out}
\label{app:generator}
The gap is not an oversight of the specification writer but a tension built into the prompt. SWE-Milestone's published generation pipeline\footnote{\url{https://github.com/DeepCommit-ai/DeepCommit}, commit \texttt{2343409d}, \texttt{harness/gen\_srs} (\texttt{srs\_gen\_v1.tmpl}, \texttt{acceptance\_test\_align.tmpl}, \texttt{prepare\_prompt.py}).} instructs the model: ``\emph{Do NOT over-specify}: No implementation hints, no file paths, no code snippets'' and to ``avoid mandating specific functions, methods, or code edits''. A second, test-aligned pass is meant to restore what tests need---``Add if tests require specific naming/interfaces: \ldots{} Module constants (when tests import them)''---and by default it analyses the new (none-to-pass) tests as well as the fail-to-pass ones. Two of its rules, however, work against the gaps we find: it tells the model to omit what the agent can ``figure \ldots{} out from the code diff'', which the agent never sees, and for a cluster of tests that share one error signature it adds ``ONE'' criterion and skips the rest. Neither pass is given a unit's existing tests, which make up \nBlockedTestsPTwoP{} of the \nBlockedUnitTests{} required tests in blocked units and fail only because a new test in their unit cannot be built. A generator told not to over-specify, and to write one criterion per error cluster, is likely to leave out exactly the names that decide whole units.

\section{Harness Design}
\label{app:harness}
SWE-Chain's harness copies the agent's tree over the reference image, so missing modules are filled in; it turns a failed import into a placeholder, so only tests that use the name fail; and it reports failed imports back to the agent for one repair round. The build report named the hidden name in \chainFeedbackReported\% of the system runs, and \chainFeedbackWritten\% of runs ended up writing it. SWE-Milestone's harness instead grafts the reference tests onto the agent's tree and fails whole build units, and gives no such feedback.

\section{Reproduced Hidden Names in Detail}
\label{app:memorization}
\textbf{Evidence.} Whether a run provided a hard dependency is read from its evaluation (the release's source snapshots, which we read only for the example of Section~\ref{sec:motivation}, would allow a direct check): a passing required test of a unit whose build needs the dependency (Go, Rust and Java build the whole unit; Python fails module-level imports; TypeScript fails unresolved module paths) shows that it existed; a log line naming it as undefined, not found or not exported shows that it did not; otherwise the (item, run) pair is undetermined. A failed test is no evidence of a build, because some runs record every test of a unit that did not compile as failed.

\textbf{Rates.} Without the run marked as leaked, runs reproduce 45.2\% of 2,056 determined (hard item, run) pairs and 79.0\% of 1,064 inferable ones. Rates differ by repository: in go-zero 34.4\% against 49.9\%; in the TypeScript front ends, where most hard items are module paths, 90--100\%. Before the path rule of Section~\ref{sec:approach}, this contrast motivated it: the paths the rule now treats as inferable were written in 99.7\% of the determined pairs (652 of 654), the paths it keeps hard in \pctPathHardWritten\% (\nPathHardWritten{} of \nPathHardDeterminate).

\textbf{Exposure.} For each run and repository we compare the vendor-stated training (or knowledge) cutoff of the model with the release dates of the repository's interval. Within repositories, runs whose cutoff falls after the interval differ from the others by \memorizationWithinDiff{} points on hard items (permutation $p$ = \memorizationWithinP); across the runs with a stated cutoff, the partial correlation between a run's exposed share and its reproduction rate, given its official score, is \memorizationPartialR{} ($p$ = \memorizationPartialP). The analysis is correlational and has little power, but it gives no sign that exposure to the upstream code explains which runs reproduce hidden dependencies.

\section{Representative Cases}
\label{app:cases}

\textbf{Tests the benchmark silently dropped (go-zero, B3).} The reference tests of M017 use names such as \texttt{maxFormParamCount} that M017's own gold state does not define, so the reference solution cannot compile them. The benchmark's classification recorded the affected tests as pass-to-none, and they disappeared from the required set. The defect lowers no score, but it removes coverage that the milestone's requirements call for.

\textbf{A milestone no run could build (nushell, order conflict).} Milestone G05\_be6e868 refactors the language-server tests, whose reference initialises \texttt{PositionalArg} without the field \texttt{custom\_completion} that milestone G01\_48bca0a adds. Agents implement both in one tree; the 19 runs that had implemented G01\_48bca0a first could not compile the grafted tests because of the missing field, the other runs stopped at other build errors in the same tests, and no published run produced a single result for the milestone's \nGFiveRequired{} required tests.

\section{Reviews of the Other Benchmarks}
\label{app:reviews}
Table~\ref{tab:reviews} gives, per benchmark, what was audited, what was flagged and how the flags were reviewed. Every review applies the definition of Section~\ref{sec:approach}: an item is confirmed when a scored test needs this exact name, module path or text, the task text does not give it, and the base neither defines it nor determines it by convention; otherwise it is a false positive. SWE-Milestone is reviewed per build unit by two blind annotators (RQ4). The other benchmarks are reviewed per flagged item against the task text (for SWE-EVO also the pull-request and issue text its agent instructions append), the tests and the reference change, with first-pass verdicts drafted by an LLM and checked by an author. Where a benchmark has more flags than we reviewed, the reviewed items are a random sample stratified by repository (SWE-EVO) or package (SWE-Chain), seed 20261010. Units, reviewers and sample sizes differ, so the precisions are not comparable across rows, and the intervals are wide where few items were reviewed: Table~\ref{tab:benchmarks} is exploratory evidence that construction matters, not a controlled comparison. Every verdict, with its quotations, is released with the data.

\begin{table}[h]
\centering\footnotesize\setlength{\tabcolsep}{1.6pt}
\begin{tabular}{@{}lrrrrr@{}}
\toprule
Benchmark & Tasks & Flag. & Rev. & Conf. & Prec. (95\% CI) \\
\midrule
SWE-Milestone & 98 & 89 & 65 & 58 & 89.2 (79.4--94.7) \\
SWE-EVO & 48 & 79 & 40 & 24 & 60.0 (44.6--73.7) \\
SWE-Chain & 155 & 357 & 40 & 28 & 70.0 (54.6--81.9) \\
SWE-bench Verif. & 500 & 4 & 4 & 3 & 75.0 (30.1--95.4) \\
DeepSWE & 113 & 23 & 23 & 1 & 4.3 (0.8--21.0) \\
\bottomrule
\end{tabular}
\caption{Review protocol per benchmark. Tasks: graded milestones, release tasks, steps, instances, tasks. Flag.: flagged build units with a hidden dependency (SWE-Milestone) or flagged items, the frame each review sampled from (SWE-EVO counts an item once per test unit, 62 distinct). Rev.: reviewed; SWE-Milestone, the units of the two blind rounds drawn after the rules were frozen (RQ4); SWE-EVO and SWE-Chain, stratified random samples; Verified and DeepSWE, all flags. Conf.: confirmed. Prec.: confirmed among reviewed (\%), Wilson interval.}
\label{tab:reviews}
\end{table}

\end{document}

%% file: figures/en/hidden-name-gap-figure.tex
\begin{figure*}[t]
  \centering
  \includegraphics[width=\textwidth]{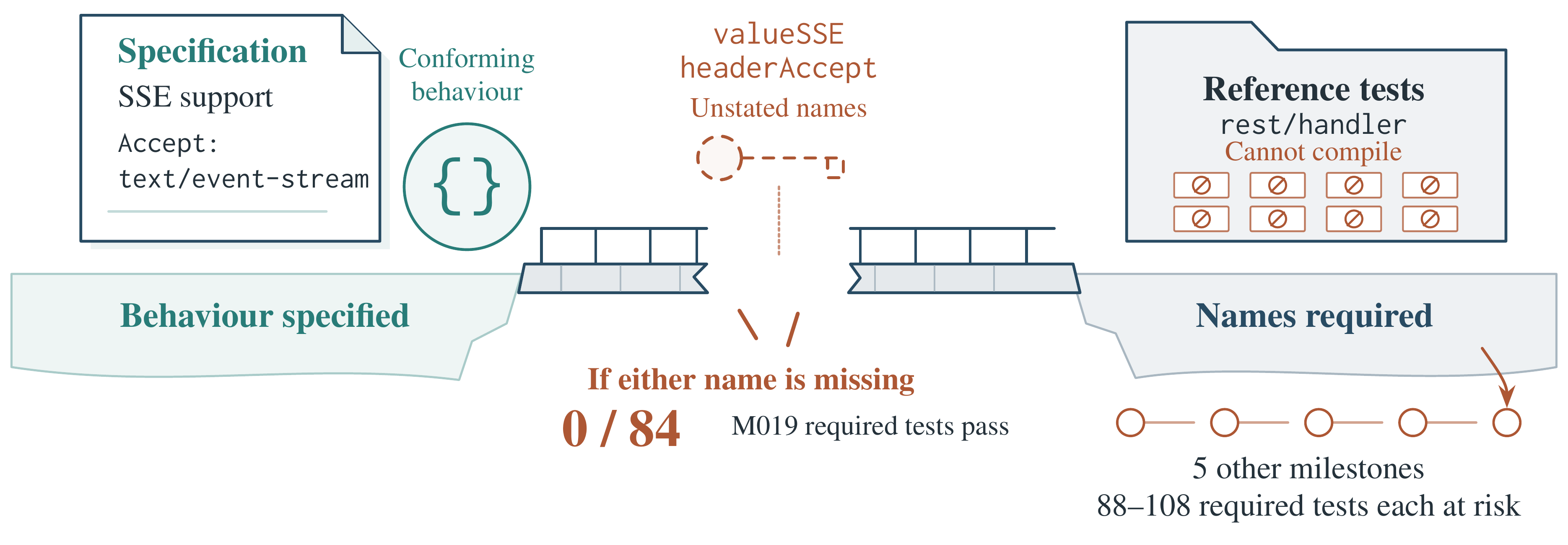}
  \caption{\textbf{Behaviour specified, names required.} The broken bridge represents an unstated interface: go-zero M019 asks for SSE behaviour but its reference tests require \texttt{valueSSE} and \texttt{headerAccept}. Missing either name blocks the whole \texttt{rest/handler} package before tests execute; the dependency recurs in five other milestones. Test tiles and milestone links are schematic.}
  \label{fig:hidden-name-gap}
\end{figure*}

%% file: figures/en/audit-overview-figure.tex
\begin{figure*}[t]
  \centering
  \includegraphics[width=\textwidth]{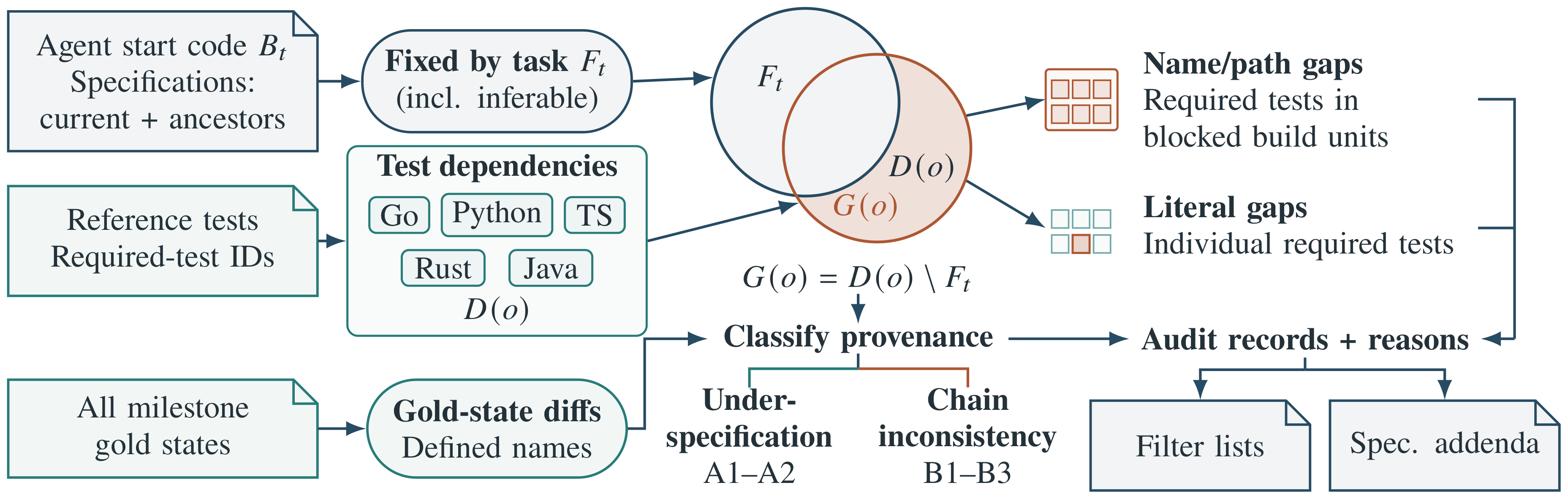}
  \caption{Static audit: five language extractors recover test dependencies (TS: TypeScript), and the orange crescent is $G(o)=D(o)\setminus F_t$, dependencies that neither the base, the specification nor inferable conventions fix. Hard name/path gaps flag all required tests in blocked build units, whereas literals flag individual tests; outputs are reasoned audit records, filter lists and specification addenda. Gold diffs distinguish under-specification (A1: current-milestone gold change; A2: ancestor gold change) from chain inconsistency (B1: non-ancestor gold change; B2: unassigned upstream commits; B3: absent from own gold); B3 tests may already be ungraded as pass-to-none.}
  \label{fig:audit-overview}
\end{figure*}